\documentclass[fleqn,10pt]{wlscirep}
\usepackage[utf8]{inputenc}
\usepackage[T1]{fontenc}
\usepackage{subcaption}

\title{Disruptive Transformation of Artworks in Master-Disciple Relationships: The Case of Ukiyo-e Artworks}

\author[1]{Honna Shinichi}
\author[2,*]{Akira Matsui}
\affil[1]{Japan Advanced Institute of Science and Technology, Co-creative Intelligence Research Area, Nomi, Ishikawa, Japan}
\affil[2]{Research Institute for Economics \& Business Administration, Kobe University, Hyogo, Japan}

\affil[*]{Corresponding author: amatsui@rieb.kobe-u.ac.jp}

\begin{abstract}
Artwork research has long relied on human sensibility and subjective judgment, but recent developments in machine learning have enabled the quantitative assessment of features that humans could not discover. In Western paintings, comprehensive analyses have been conducted from various perspectives in conjunction with large databases, but such extensive analysis has not been sufficiently conducted for Eastern paintings. Then, we focus on Ukiyo-e, a traditional Japanese art form, as a case study of Eastern paintings, and conduct a quantitative analysis of creativity in works of art using 11,000 high-resolution images. This involves using the concept of calculating creativity from networks to analyze both the creativity of the artwork and that of the artists. As a result, In terms of Ukiyo-e as a whole, it was found that the creativity of its appearance has declined with the maturation of culture, but in terms of style, it has become more segmented with the maturation of culture and has maintained a high level of creativity. This not only provides new insights into the study of Ukiyo-e but also shows how Ukiyo-e has evolved within the ongoing cultural history, playing a culturally significant role in the analysis of Eastern art. 
\end{abstract}
\begin{document}

\flushbottom
\maketitle

\thispagestyle{empty}

\section{Introduction}

Art serves significant cultural and historical roles in our society and artists mutually influence each other~\cite{vasan2022quantifying}. While recent research reveals connections among contemporary artists~\cite{vasan2022quantifying,balduf2022dude}, we understand little about the mutual relationships among historical artists in their creation of artworks in a quantitative way. Art historians qualitatively analyze and interpret the value and meaning of artworks based on style, theme, technique, and cultural context, leading to subjective evaluations influenced by their professional experiences~\cite{visualartappreciation2022}. Most research relies on historians' tacit knowledge, focusing mainly on individual works or specific artists rather than the history of art as a whole~\cite{Legitimatejudgmentart2013}. While evaluations and research by art historians significantly contribute to our knowledge, the reliance on qualitative measurements results in a lack of longitudinal analysis from massive historical records.

Recently, researchers have been conducting qualitative evaluation of art~\cite{cetinic2022understanding} using machine learning models powered by digital technologies~\cite{cardinali2019digital,valencia2024using}, assessing creativity from images of artworks~\cite{elgammal2015quantifying, spee2023machine, creativityArts2019, pelowski2017creativity}. For example, Convolutional Neural Network (CNN) can identify unique brushstrokes of different painters attribution~\cite{ji2021discerning}, and image recognition techniques are also popular~\cite{ImageNet2015,He2015,PASCALVOC}. Such studies learn the characteristics of the works for applications in art historical research, cataloging, and even detecting forgeries~\cite{Discerningpainterhand2021, ComputationalFormalism2023}.
In other qualitative research on artworks with machine learning, researchers can detect compositional patterns that serve as artist or style signatures and reveal non-obvious historical groupings through network analysis~\cite{Lee2020Dissecting}.
Not only CNN, but also an advanced multimodal AI (Stable Diffusion), can analyze 500 years of Western painting and reveal the importance of contextual information over formal elements retrieved from painting images~\cite{Kim2025ContextAware}.
Those advancement of technology and data availability enables us to conduct large-scale quantitative analysis, for example using image recognition techniques~\cite{ImageNet2015,He2015,PASCALVOC}.

Thanks to such unsupervised models we uncover the hidden structure or mechanism from the artwork in a data-drive manner~\cite{spee2023machine,liu2021understanding}, but the evolution of artwork and artists' mutual interdependence lacks transparency. There is, for instance, a universal pattern of artists' performance within their career trajectory~\cite{liu2021understanding}. Additionally, we know little about how mutual influence drives creativity. While we understand the impact of mutual influence on creativity in scientific teams~\cite{zeng2022impactful,zeng2021fresh}, this understanding does not fully translate to the realm of artistic creativity. While the connections between paintings can estimate creativity of artworks~\cite{elgammal2015quantifying} and their diversity patterns linked to emerging styles~\cite{Kim2024Diversity}, we need to have a model that tract the trajectory of artists creativity extracted from their connections thorough artworks. 

To address the aforementioned gap in the quantitative analysis of artworks, we propose a machine learning-based approach to studying the influence of an artist's master-disciple relationship on their work style. We construct an artist network that captures the extent to which created artworks are influenced by past works, and we calculate the trajectory of disruption~\cite{park2023papers} over 100 years to study their creativity. We apply the proposed framework to the historical Japanese Ukiyo-e artwork dataset, enabling the analysis of long-term artist influences. The large number of widely appreciated artworks, the period of cultural isolation, and the strong master-disciple system provide a clear basis for examining mutual artistic influences\cite{goldberg1977ukiyo, gelunas2004making, makino1995reversion}.

\begin{figure*}[!htb]
\centering
\includegraphics[width=0.8\textwidth]{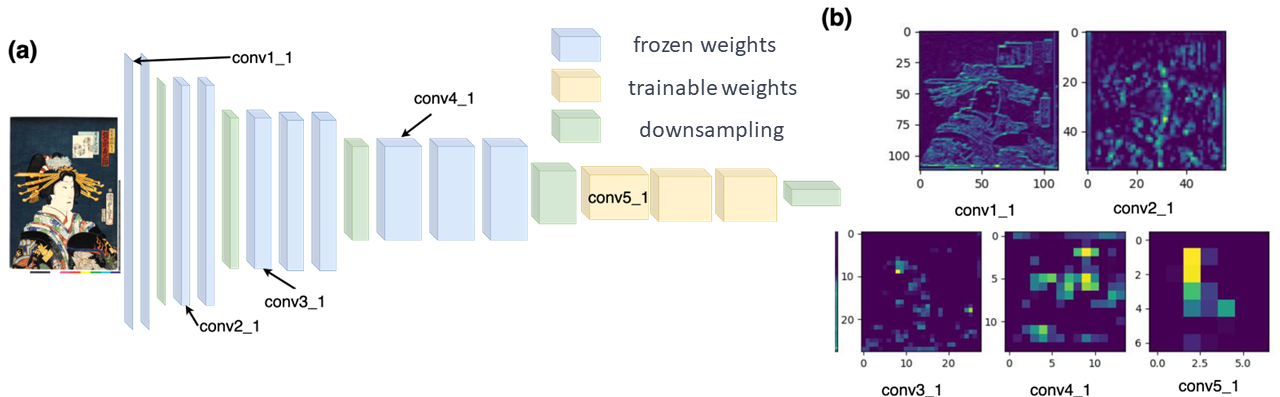}
\caption{Conceptual Diagram of the Model Used}
\label{fig:model_summary}
  \begin{minipage}{1\columnwidth}
    { { (a) The figure schematically shows a fine-tuned VGG model. The part with frozen weights has fixed weights, while the part with trainable weights is being trained. The model classifies ukiyo-e images by determining which ukiyo-e artist created them when we input an ukiyo-e image. (b) The features obtained from the layers within the model are displayed as a heatmap. High-dimensional features extract linear characteristics, while low-dimensional features capture the differences in painting styles specific to each artist.} 
    }
  \end{minipage}

\end{figure*}

\section{Data and Methods}

\subsection{Dataset}
\begin{figure*}[!htb]
\centering
\includegraphics[width=0.8\textwidth]{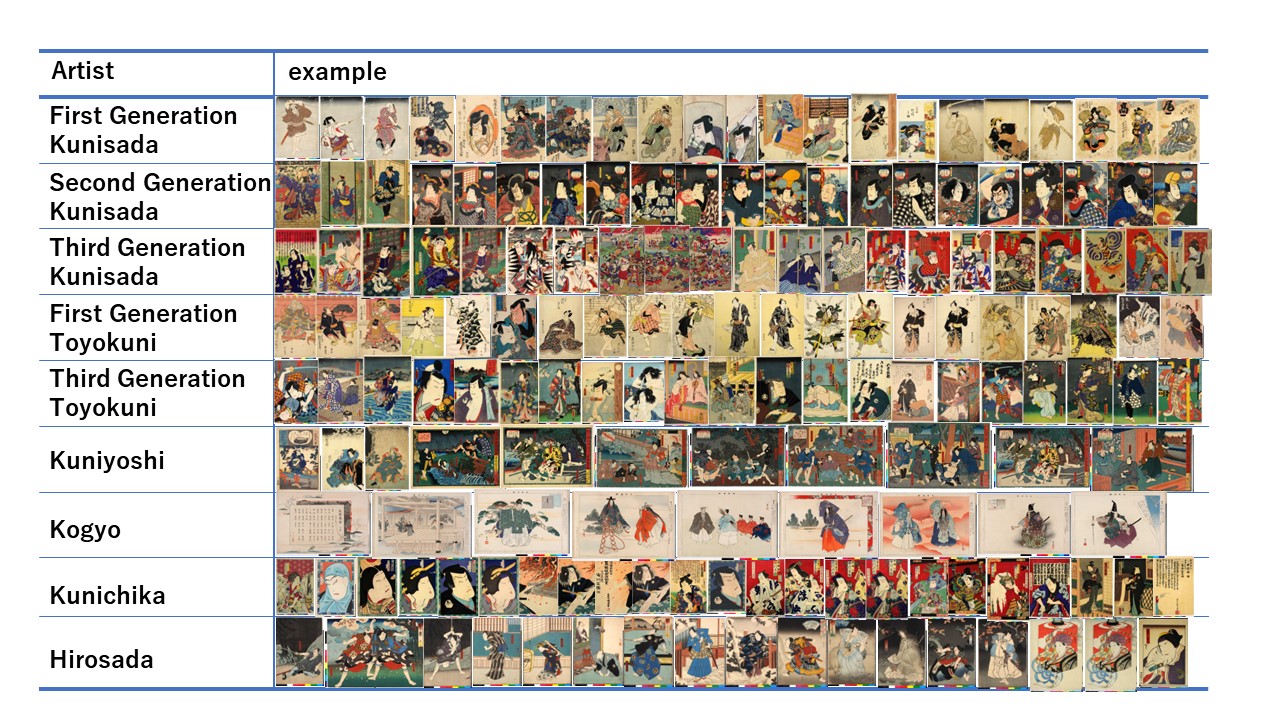}
\caption{Ukiyo-e Faces Dataset}
\label{fig:ukiyoe_face_data}
  \begin{minipage}{1\columnwidth}
    { {Ukiyo-e paintings associated with their authors as well as distribution of years with respect to these authors in our dataset. We show here nine authors with the most painting in our dataset.} 
    }
  \end{minipage}
\end{figure*}

This research learns the embedding of Ukiyo-e artwork from the ``ARC Ukiyo-e Face Dataset`` ~\cite{tian2021ukiyoe}. This dataset contains information such as ID, title, production era, author name, and images of Ukiyo-e for each artwork. The central aim of the dataset is to provide portraits of people and therefore does not include landscape paintings. It contains about 11,000 images and corresponding metadata.  Among the painters in this dataset, those who have more than 100 Ukiyo-e prints in the dataset are Kunisada, Kunisada, Kunisada, Toyokuni, Toyokuni, Kuniyoshi, Kogyo, Kunichika, and Hirosada (Table~\ref{table:num_images}). We plot the sample of those artists from the dataset in Figure ~\ref{fig:ukiyoe_face_data}. These Ukiyo-e artists have more than 100 pictures in the dataset that have all three pieces of information (image, artist name, and created year).

\begin{figure*}[!htb]
\centering
\includegraphics[width=0.8\textwidth]{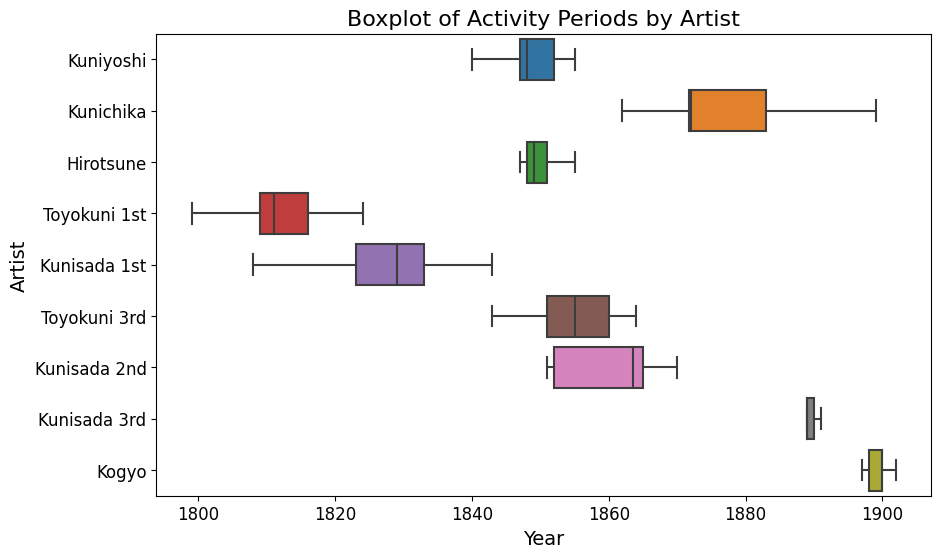}
\caption{The active era of Ukiyo-e artists.}
\label{fig:ukiyoe_era}
  \begin{minipage}{1\columnwidth}
    { Ukiyo-e paintings associated with their authors as well as distribution of years with respect to these authors in our dataset. We show here ten authors with the most painting in our dataset.} 
  \end{minipage}
\end{figure*}

\begin{figure*}[!tb]
\begin{minipage}{0.5\textwidth}
  \centering
  \includegraphics[width=1.0\linewidth]{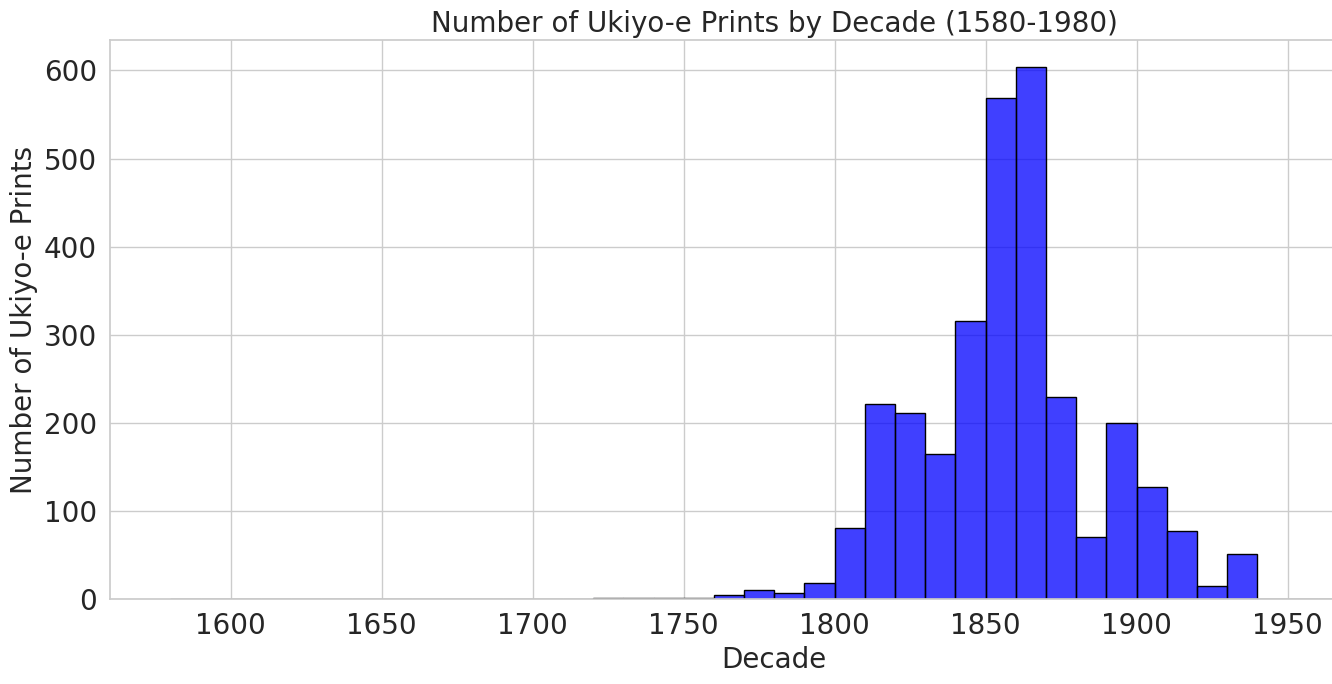}
  \caption{Number of Ukiyo-e Prints by Decade (1580-1980)}
  \label{fig:num_of_ukiyoe_by_decade}
\end{minipage}
\begin{minipage}{0.5\textwidth}
  \centering
  \includegraphics[width=1.0\linewidth]{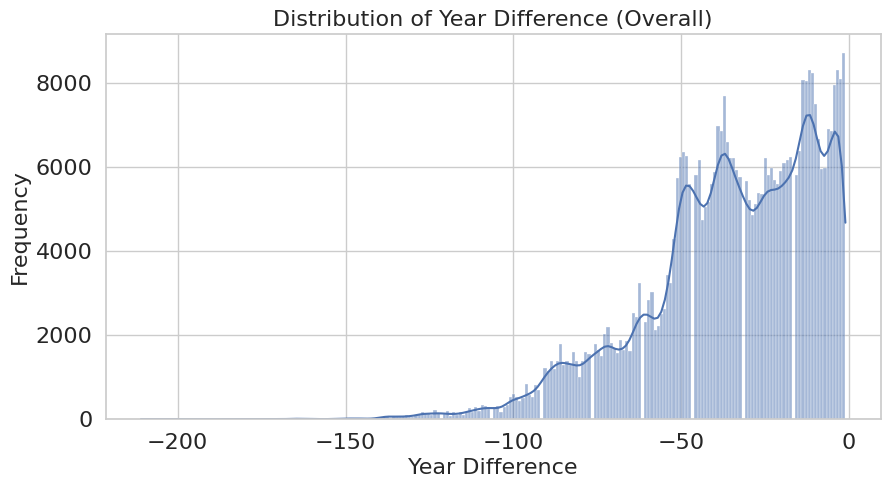}
  \caption{Distribution of year Difference}
  \label{fig:Distribution_of_year_Difference}
\end{minipage}
  \begin{minipage}{1\columnwidth}
    { {These figures show the number of Ukiyo-e produced per decade and the extent to which there are age differences in the Ukiyo-e connected between networks. Ukiyo-e production has been declining since the Meiji Restoration of 1868. In addition, the frequency of being connected in the network has decreased after 50 years in the network.} 
    }
  \end{minipage}
\end{figure*}

In this study, we obtain the characteristics of the painting style from the deep learning model by classifying the representative Ukiyo-e artists who painted many Ukiyo-e works. The reason for using this method is that there is no annotation data for each painter. In Western painting, there are painting styles such as Impressionism, but in Ukiyo-e, painting styles are not explicitly assigned to each Ukiyo-e artist.  
Figure\ref{fig:ukiyoe_era} indicates the age of each artist's activity. For each artist, a box-and-whisker chart was used to visualize which period of roughly 100 years the artist was active, from Toyokuni I to Kogyo. And, Figure \ref{fig:num_of_ukiyoe_by_decade} shows that the most Ukiyo-e artists were active between 1850 and 1860, and the number of Ukiyo-e works during this period is the largest for Ukiyo-e as a whole.

\begin{table}[ht]
\centering \small
\caption{Summary of Ukiyo-e Artists and Image Counts}
\label{table:num_images}
\begin{tabular}{clll}

\textbf{ID} & \textbf{Ukiyo-e Artist (Active Period)} & \textbf{Number of Images Used} &  \textbf{Generation (if applicable)}\\

\hline\hline
1 & Kunisada (1808--1843) & 245 & 1st \\
2 & Kunisada (1851--1870) & 146 &2nd\\
3 & Kunisada (1872--1903) & 101 & 3rd\\
4 & Toyokuni (1793--1827) & 222 & 1st\\
5 & Toyokuni (1844--1864) & 673 & 3rd\\
6 & Kuniyoshi (1825--1855) & 164 & \\
7 & Kogyo (1897--1924) & 208 & \\
8 & Kunichika (1862--1899) & 344 & \\
9 & Hirosada (1847--1855) & 165 & 
\end{tabular}
  \begin{minipage}{1\columnwidth}
    { {This table shows the number of Ukiyo-e artists and their Ukiyo-e works. The model adjusted the data based on the number of Ukiyo-e artists with more than 100 pictures, taking into account the training costs.} 
    }
  \end{minipage}
\end{table}


\subsection{Extracting latent features from Ukiyo-e artworks}\label{sec:feature_vgg}

To extract features from images, we employ Very Deep Convolutional Networks (VGGNet), to where layered architecture consists of multiple convolutional and fully connected layers with small filter sizes~\cite{simonyan2014very}. Building on these attributes, the ``A Neural Algorithm of Artistic Style`` research highlights the network's distinct capabilities in artistic applications~\cite{gatys2015neural}. The approach utilizes the network’s varying layers for feature extraction, skillfully differentiating and transforming style and content in images. The network's layered structure facilitates detailed feature extraction, with earlier layers handling finer content details and deeper layers capturing more abstract style elements. We use this property to obtain features from each layer when extracting its style from an image. The initial layers of this model generally capture basic structural features that are universal across images, while deeper layers are more adept at identifying features specific to the image dataset in question~\cite{jogin2018feature}. By fine-tuning the features in the fifth layer of the style network, we obtain the style of Ukiyo-e in a manner that accentuates the differences between the artists. 

We present the schematic of this study in Figure~\ref{fig:model_summary}. Figure~\ref{fig:model_summary}(a) depicts the architecture of the VGGnet for the method where each rectangular block represents a convolutional or fully connected layer. The yellow segments indicate the layers that we re-trained during the fine-tuning process, the blue segments represent the layers of the fixed pre-trained VGG model, and the green segments denote pooling layers that aggregate features. As depicted in Figure~\ref{fig:model_summary}(b), each layer is responsible for capturing different features of the Ukiyo-e images, from basic textures and patterns to more complex structures and details~\cite{jogin2018feature}. To obtain the representations of images, we integrate both high and low-dimensional features (Figure~\ref{fig:model_summary}(a)~\texttt{Block1, Block2, Block5}), which are each 100 dimension vectors. We, then, reduce the dimensions of combined features into 100, using PCA~\cite{wold1987principal}. This method follows the approach used in research to estimate the period when a painter was producing his or her best work from an image~\cite{liu2021understanding}.

\subsection{Constructing Ukiyo-e Networks}\label{sec:artwork_network}~\label{sec:network_construction}

We construct the networks of ukiyo-e artworks and artists based on the extracted features of artwork images (Sect.~\ref{sec:feature_vgg}) To model the complex relationships and creative influences among Ukiyo-e artists, we constructed two types of networks that represent different perspectives on artistic influence: an \textbf{Image Similarity-Based Network (ISN)} and a \textbf{Style/School-Based Network (SSN)}. Each network captures unique characteristics of the Ukiyo-e artworks, allowing us to explore how stylistic elements propagate among artists over time.

\subsubsection{Image Similarity-Based Network (ISN)}

\begin{figure*}[!htb]
\centering
\includegraphics[width=0.8\textwidth]{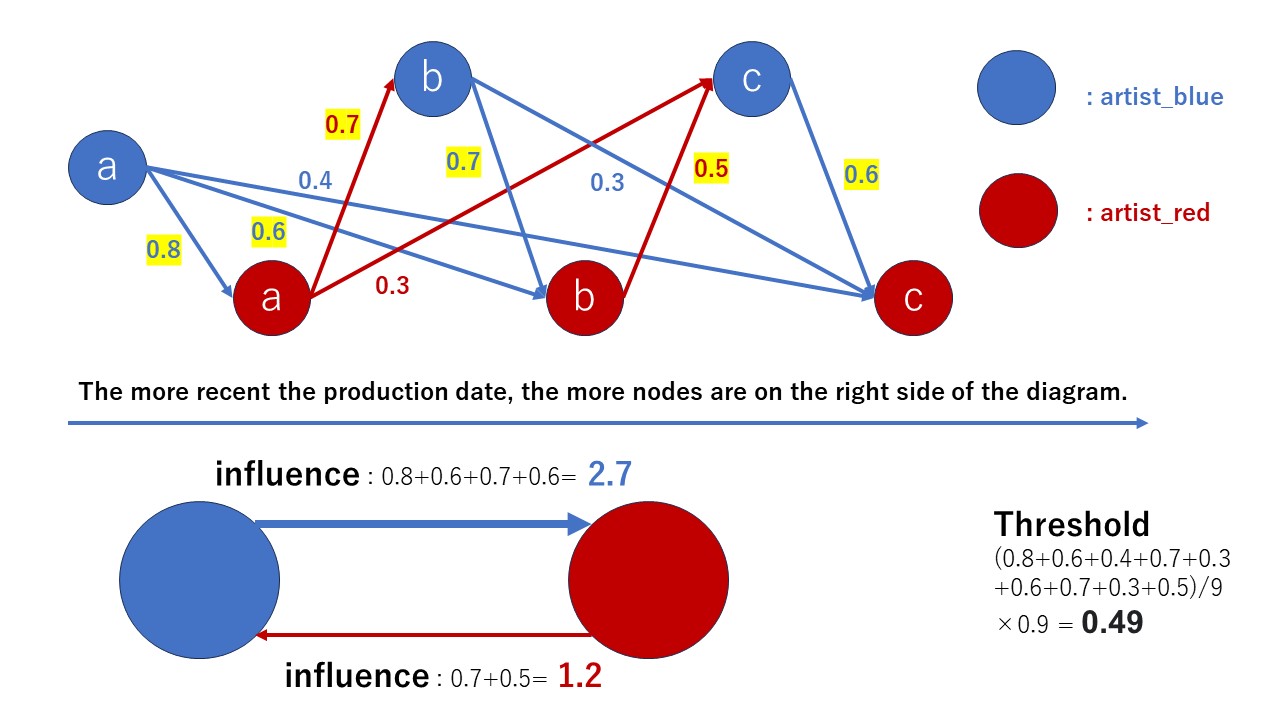}
\caption{Constructing Style-Based Network (SSN)}
\label{fig:cos_network_exp}
  \begin{minipage}{1\columnwidth}
    { {Schematically diagram of the Blue Artists and Red Artists in connecting their networks. Each node represents a Ukiyo-e artwork.} 
    }
  \end{minipage}
\end{figure*}

The ISN focuses on the visual relationships between individual Ukiyo-e artworks, where nodes represent artwork and edges represent influence (Fig.~\ref{fig:cos_network_exp}). The edges of ISN are directed that describe chronological order. To construct this network, we calculated the cosine similarity between feature vectors extracted from the images. For each artwork, edges are established to connect it with other artworks that have a high cosine similarity score, representing direct visual influence. In this way, the ISN captures the propagation of specific visual features across different works, reflecting how individual artworks might have visually influenced subsequent creations. We set a threshold to exclude edges with lower similarity scores to focus on significant relationships and ensure that the connections primarily reflect substantial stylistic influence. We calculate the cosine similarity of each painting with all other paintings to find meaningful relationships and we set the threshold as the top 90\% tile of the all distribution. Note that edges are only drawn between works by different artists to avoid bias in depicting intra-artist influence. This network construction thus emphasizes the image-level similarities, depicting how visual elements of Ukiyo-e evolved across different artists and generations.

The underlying assumption of the ISN is that an artist's paintings share similar features, making it unnecessary to track the feature propagation between an artist's own works, guiding this decision. The cosine similarity serves to measure the distance between nodes. The assumption here relies on the idea that an artist's paintings inherently possess similar features. This similarity within an artist's body of work renders the observation of feature propagation between their own paintings unnecessary for the study of influence and relationships between different artists. 

\subsubsection{Style-Based Network (SSN)}~\label{sec:method_cbn}~\label{sec:method_ssn}

While ISN captures the visual differences between artworks, it focuses on pairwise comparisons and does not consider the collective style formed by multiple artworks. To model the stylistic influence between Ukiyo-e artwork, we construct Style-Based Network (SSN). The SSN emphasizes the stylistic relationships within established schools or styles in the Ukiyo-e tradition. In the construction of SSN, we first group artworks into clusters based on their stylistic features using K-means clustering, assigning each artwork to a specific cluster that represents a particular style or thematic category.

For SSN, we follow a procedure similar to ISN. In SSN, nodes represent artworks and edges represent influence, but we focus on influence within a style.
To identify the styles of Ukiyo-e through cluster classification, we first group the artworks by K-means clustering~\cite{hartigan1979algorithm} and obtain 20 clusters. We further consolidate any cluster containing three or fewer paintings into an ``other'' category, resulting in a classification that identifies six unique types of clusters and the ``other'' cluster. Thanks to the ``other'' cluster, we identify meaningful groups for our analysis without focusing on the optimal number of clusters, a controversial issue~\cite{Ikotun2022Kmeans}.

\begin{figure}[!htb]
\centering
\includegraphics[width=0.8\textwidth]{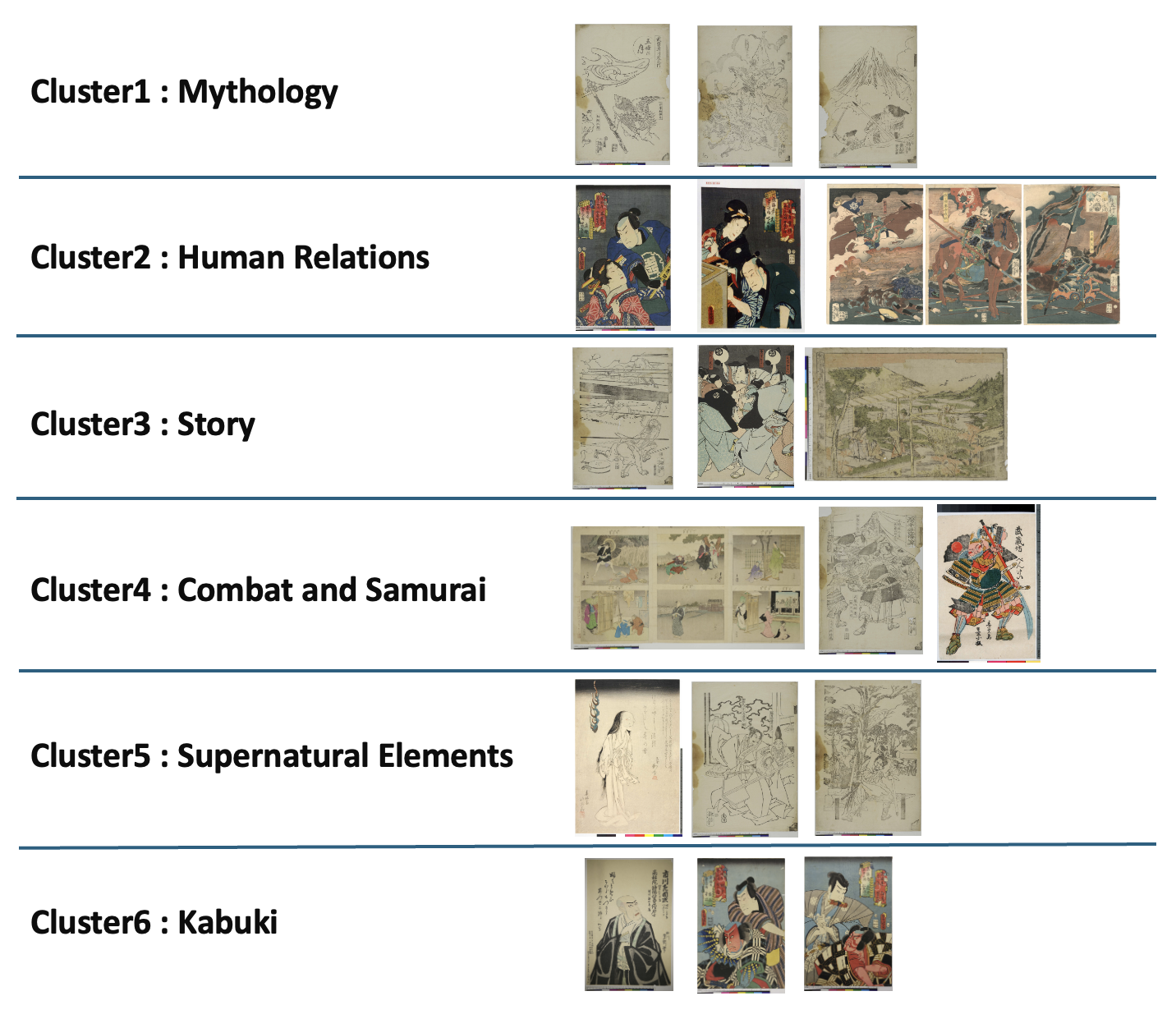}
\caption{Style-Based Clusters}
\label{fig:cluster_mean}
  \begin{minipage}{1\columnwidth}
    { {Clusters of artworks based on features obtained from the VGG model. The presented images show the centers of each cluster.} 
    }
  \end{minipage}
\end{figure}

For each cluster, we select the three Ukiyo-e paintings that are closest to the cluster's center. This approach allows us to observe and analyze the style that each cluster represents, providing insights into the diverse characteristics of Ukiyo-e art. Cluster 1 includes themes related to mythology and legends, featuring distinctive compositions. Cluster 2 depicts multiple people, showing interactions among them. Cluster 3 focuses on episodes or stories, featuring scenes from these narratives. Cluster 4 centers on battles and samurais. Cluster 5 explores themes of ghosts and the supernatural. Cluster 6 showcases Kabuki actors, capturing their characteristic expressions. We show the images for each cluster in Figure~\ref{fig:cluster_mean}.

\begin{figure}[!htb]
\centering
\includegraphics[width=0.8\textwidth]{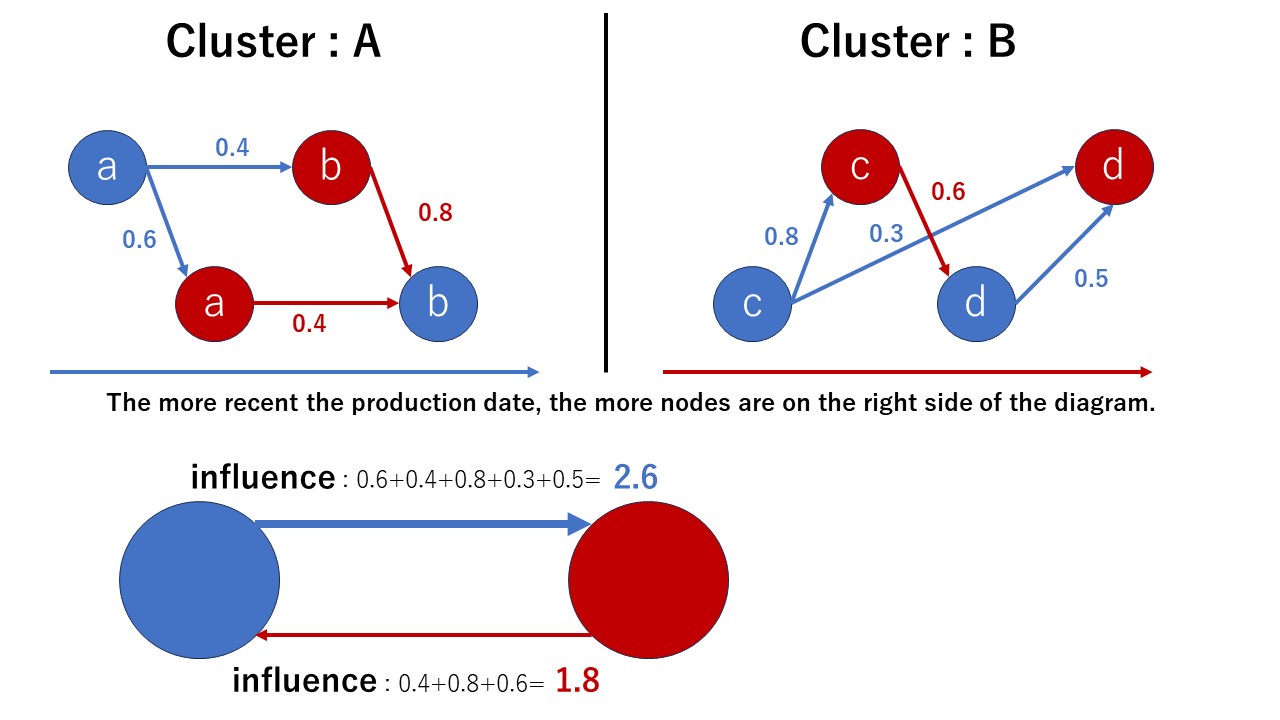}
\caption{Constructing Clustering Based Network (CBN)}
\label{fig:cluster_network_exp}
  \begin{minipage}{1\columnwidth}
    { {This diagram represents the network of blue and red artists, where each node denotes one Ukiyo-e. We classify each Ukiyo-e into Cluster A or Cluster B and determine the influence values within each cluster using cosine similarity.} 
    }
  \end{minipage}
\end{figure}

After obtaining the clusters of artwork style, we connect the nodes within the same cluster in chronological order, taking edges from old to new within that cluster. To consider the strength of influence within the same cluster, we represent the distance of the edges by cosine similarity, observing how much one painting directly influences another. We create the network between Ukiyo-e artists in a similar way to the CSN. Figure~\ref{fig:cluster_network_exp} schematically represents this network structure.

This network provides an important perspective for understanding how artists influence each other within a certain style because it accounts for the style comprehensively determined from the entire set of images through clustering.

\subsubsection{Projecting artwork networks into artist network}
The purpose of the ISN and SSN is to observe how Ukiyo-e has changed in terms of picture similarity by constructing an entire network from relationships between individual pictures. At the same time, we construct a network that maps the relationships between Ukiyo-e artists based on the connections among their artworks. In this network, we translate the connections between individual Ukiyo-e pieces into direct relationships between the artists themselves. Thus, the link from one Ukiyo-e to another reflects the connection between the artists who created those pieces, forming a network of artist relationships derived from the network of their artworks. We carry out this process for both the CSN and the CBN.

In the artists networks, the size of the edge connecting Ukiyo-e artists represents the degree of influence of the Ukiyo-e connected by the edge, or the number of Ukiyo-e connected by the edge. For example, suppose there are artists \(A\) and \(B\), and they have Ukiyo-e \((a_1, a_2)\) and \((b_1, b_2)\) respectively. Additionally, assume all of \(A\)'s works are older than \(B\)'s. If the degree of influence between \((a_1)\) and \((b_1)\) is 1, and the degree of influence between \((a_1)\) and \((b_2)\) is 0.5, and there are no edges connecting \((a_2)\) with \((b_1)\) or \((b_2)\), then the degree of influence between artist \(A\) and artist \(B\) is \(1 + 0.5 = 1.5\).

\subsection{Index Calculated on the Networks}
This section describes the indices calculated on the constructed networks. We employ the betweenness centrality measurement and the Disruption Index (D5).

\subsection{Betweenness Centrality}

We analyze the networks we have created from two perspectives: how crucial the roles of nodes are for network propagation, and how much they incorporate and develop new features. By examining betweenness centrality and the disruptive index, we observe how new characteristics of Ukiyo-e emerge and spread. Betweenness centrality quantifies the number of times a node acts as a bridge in the shortest path between two other nodes~\cite{barthelemy2022betweenness}. This metric reveals an artist's influence in the network, including the potential to control and facilitate communication flows. A higher value indicates a stronger contribution to the propagation of features within the network, reinforcing its significance.

\begin{equation}
    C_B(v) = \sum_{s \neq v \neq t} \frac{\sigma_{st}(v)}{\sigma_{st}},
\end{equation}
where \(v\) is a node in the network, \(\sigma_{st}\) is the total number of shortest paths from node \(s\) to node \(t\), and \(\sigma_{st}(v)\) is the number of those paths that pass through \(v\).

\subsubsection{Disruption Index}\label{sec:method_d5}
Betweenness centrality measures the importance of a node, but it does not capture how novel the node is. Therefore, we use the disruption index, which evaluates a node's novelty and influence. The disruptive index originated in the field of Science of Science~\cite{zeng2023disruptive} to quantify how novel and influential a paper is in terms of its citation relationships.

\begin{figure}[!htb]
\centering
\includegraphics[width=0.8\textwidth]{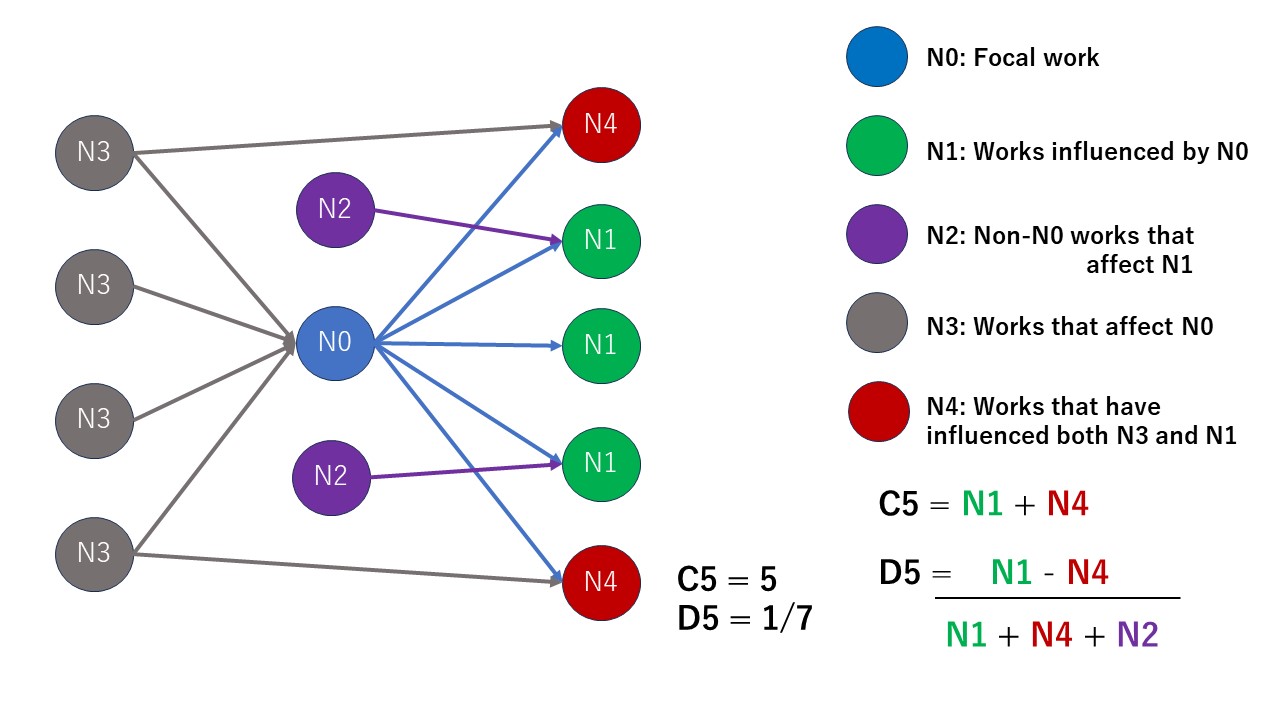}
\caption{Calculating Disruption Index of Ukiyo-e Networks}
\label{fig:D5_exp}
  \begin{minipage}{1\columnwidth}
    { {The schematic illustration of Disruption Index (D5) calculation. The index of N0 captures how much N0 influences others, considering its uniqueness.} 
    }
  \end{minipage}
\end{figure}

Figure~\ref{fig:D5_exp} illustrates the Disruption Index (D5) on the networks. Consider the calculation of the index of node N0. This process involves nodes influencing N0 (nodes extending edges to N0), and nodes influenced by N0 (nodes to which N0 extends edges). Let N1 represent nodes influenced by N0, and let N2 represent nodes influencing N1 that are not N0. Nodes influencing N0 are denoted as N3, and nodes among N1 also influenced by N3 are N4. The influence count C5 equals the sum of N1 and N4. To calculate the disruptive index D5, subtract N4 from N1, and then divide by the sum of N1, N4, and N2. C5 indicates the total number of paintings influenced by N0, while D5 uses the difference between N1 and N4 as its numerator. This difference reflects the disruptive features that N0 newly generated, and the denominator normalizes this value with the paintings influenced by N0 and by others on N1. Thus, N0's disruptive features do not inflate solely due to its extensive influence or strong influence from others.

\section{Result}
We construct the networks that describe the propagation of artistic style from the embedding model learned from the Ukiyo-e art dataset, which captures the stylistic influence on Ukiyo-e artists. With the constructed networks, we first demonstrate that the Ukiyo-e became less disruptive over time in terms of the similarity of Ukiyo-e artworks. On the other hand, our analysis reveals that the disruptiveness within each school of Ukiyo-e did not decline and remained stable throughout the period. In addition, our analysis detects the influential individuals whose styles subsequent artists followed, and identified the ``masterpiece'' in terms of its impact on subsequent artistic styles.

\subsection{Disruptiveness Analysis of Ukiyo-e Artists}\label{sec:artwork_disruptiveness}

We calculate the trajectories of the disruptiveness from the two different networks that capture different natures of the relationships between artworks. Our first network is the Cosine Similarity Network (CSN), which emphasizes the relationships between individual artworks, rather than examining the positioning of these works within the broader context of Ukiyo-e works (i.e., the perspective of artistic schools). To capture the schools of style, we construct the Clustering Based Network (CBN) where similar artworks in the same clusters are connected based on their order of publications (For the method of network construction, see Method Sec.~\ref{sec:network_construction}). The two types of networks allow us to study the disruptiveness of the Ukiyo-e artworks from two different perspectives; CSN captures the role of individual artworks; CBN captures the evolution of artworks within its schools. 

\begin{figure}[!t]  
  \centering
  \begin{subfigure}{0.495\textwidth}
    \centering
    \includegraphics[width=1.0\linewidth]{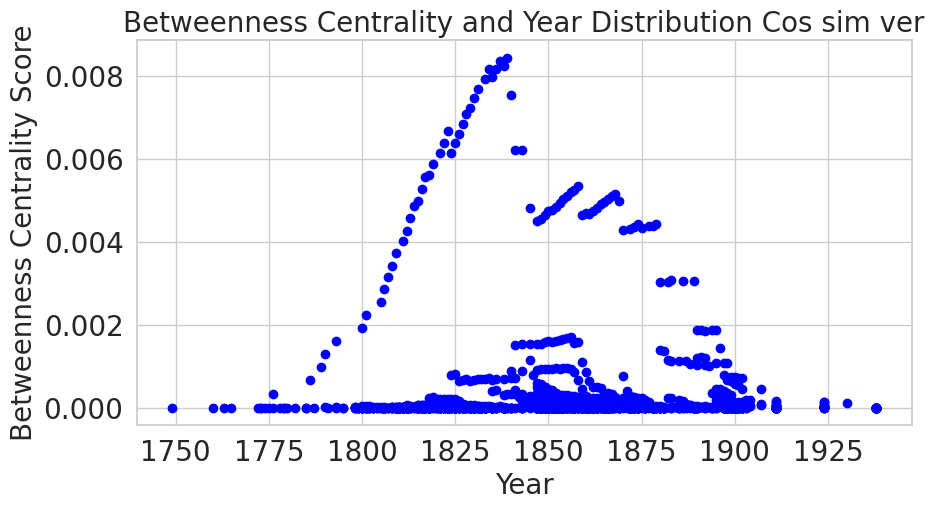}
    \caption{Cosine Similarity Network (CSN)}
    \label{fig:cosine_betweenness}
  \end{subfigure}
  \hfill
  \begin{subfigure}{0.495\textwidth}
    \centering
    \includegraphics[width=1.0\linewidth]{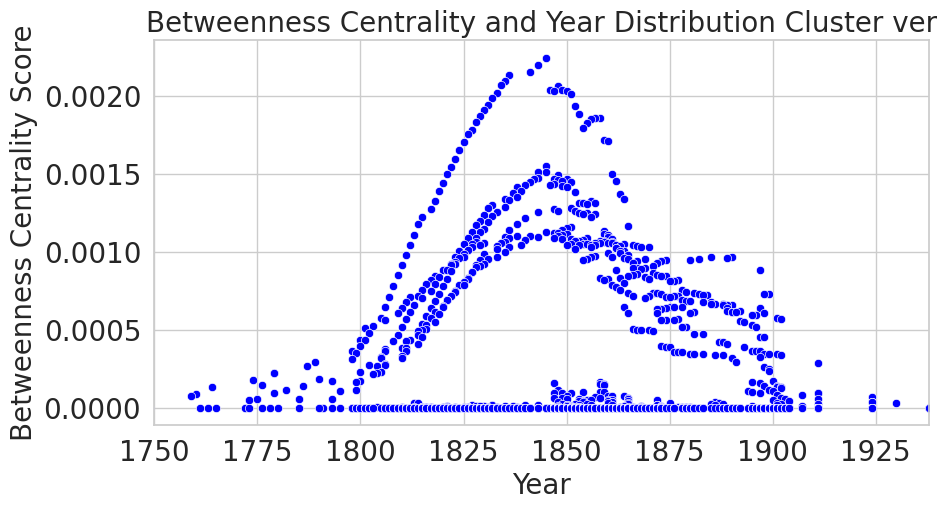}
    \caption{Clustering Based Network (CBN)}
    \label{fig:cluster_betweenness}
  \end{subfigure}
\caption{Trajectories of the betweenness of Ukiyo-e artworks over the period}\label{fig:betweeness_csn_cbn}
  \begin{minipage}{1\textwidth}
    \raggedright %
    { The distribution of Betweenness Centrality per Ukiyo-e for CSN and CBN is shown. Both are mountainous, but the position of the vertex and the shape of the mountain are different.}
  \end{minipage}
\end{figure}

To briefly depict the properties of the constructed networks, we calculate the betweenness centrality of nodes in both networks. Since the nodes represent artworks in the two networks, such indicators can capture the importance in terms of connectivity in the networks. We plot the trajectory of the betweenness throughout data periods in Figure~\ref{fig:betweeness_csn_cbn}.

Both CSN and CBN reveal that they reached their peak around the 1840s and experienced a decline around the 1850s-60s, during which several historical events influenced the networks. First, that period saw technological innovations, such as the introduction of synthetic dyestuffs~\cite{cesaratto2018timeline}. Second, the opening of the country~\cite{bell2004explaining} and the influx of Western paintings had a significant impact on Ukiyo-e. For example, it is known that the influx of Western realism and vibrant colors introduced during the Meiji Restoration likely led to the inclusion of motifs reflecting modernization and Western influences in paintings~\cite{kafu2012ukiyo}. It should be mentioned that this historical event gradually transformed the style of Ukiyo-e, though the pace was slower than other rapid changes in Japanese society, such as in the economy or social systems. For example, \cite{cesaratto2018timeline} points out that the introduction of synthetic dyes was not a rapid transformation but rather a carefully selected and gradual process.

The trajectories of betweenness centrality in CSN and CBN also indicate that the decline of CBN is slower than that of CSN. This gradual decline might suggest that adapting to new painting styles can be a lengthy process, with its influence fading more slowly. However, some elements of the Ukiyo-e painting style possibly persisted to some extent within these new styles. Nonetheless, as the cultural focus gradually shifted towards new styles represented by synthetic dyes and other products with exceptionally bright colors and consistency introduced during the Meiji era, the scores of betweenness centrality gradually decreased.

\begin{figure}[!htb]  
  \centering
  \begin{subfigure}{0.45\textwidth}
    \centering
    \includegraphics[width=1.0\linewidth]{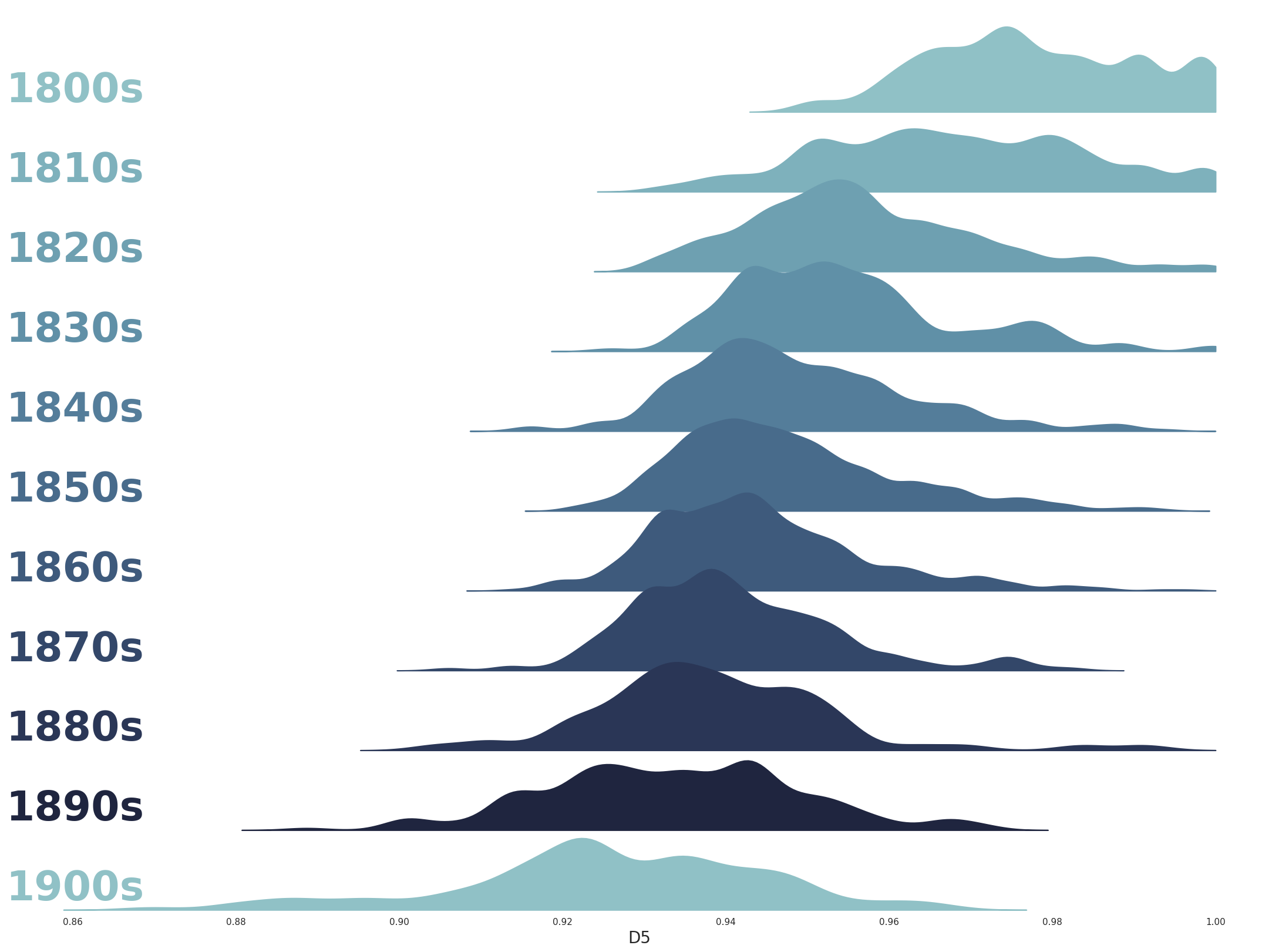}
    \caption{Cosine Similarity Network (CSN)}
    \label{fig:trajectory_cos_D5}
  \end{subfigure}
  \hfill
  \begin{subfigure}{0.45\textwidth}
    \centering
    \includegraphics[width=1.0\linewidth]{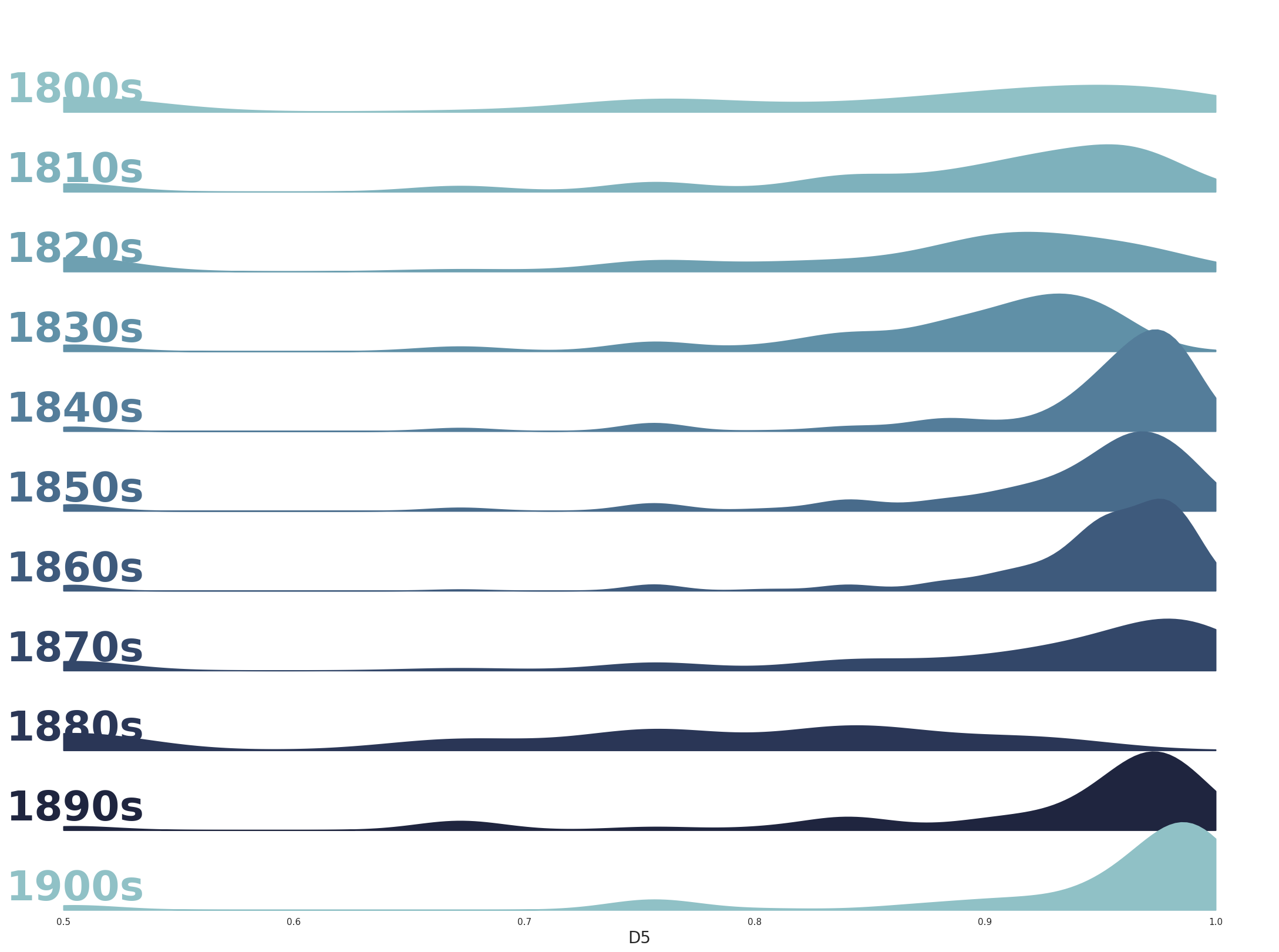}
    \caption{Clustering Based Network (CBN)}
    \label{fig:trajectory_cluster_D5}
  \end{subfigure}
\caption{Trajectory of disruption}
  \begin{minipage}{1\textwidth}
    \raggedright %
    { Decadal histograms of the distribution of desruption index (D5) for each Ukiyo-e in the CSN and CBN; the destructibility indices in the CSN show a decreasing trend with each passing year, while in the CBN they remain stable at high values.}
  \end{minipage}
\end{figure}

While the aforementioned analysis of betweenness centralities captures the trajectories of historical artworks, this simplified approach may overlook critical aspects of the relationships such as their significance in terms of creativity. To understand the propagation of creativity through artistic works on the network, we calculate the disruption index (D5; Method Sec.~\ref{sec:method_d5}) and track its trajectory over the course of each decade. We first calculate D5 index of CSN and present in Figure~\ref{fig:trajectory_cos_D5}. We find that the disruptiveness of works diminishes over the data period, consistent with the dynamics observed in the analysis of scientific works in which the disruptiveness index was originally proposed~\cite{zeng2023disruptive}. The figure shows elevated D5 values in the initial period (1800s--10s) with substantial variance, characterizing the long tail of the distribution. For example, one of the most famous works of the 1830s is {\it Thirty-six Views of Mount Fuji}~\cite{wikipedia_mount_fuji}, recognized as a masterpiece of Ukiyo-e. During the middle period, our analysis reveals a bell-shaped distribution. Finally, the figure exhibits a downward trend in the average D5 values in the later periods. This analysis with CSN indicates a declining trend of the D5 index, suggesting that the Ukiyo-e artwork became less disruptive during the data period. This
result tempts us to consider that artistic work might experience a similar phenomenon as observed in scientific work~\cite{zeng2023disruptive} that the decline of disruptive work as its domain matures.

On the other hand, the dynamics of D5 calculated for CBN do not reveal a clear trend, which studies the relationships among artworks from different angles and is plotted in Figure~\ref{fig:trajectory_cluster_D5}. The CBN focuses on the relationships between artworks within its school (Method Sec~\ref{sec:method_cbn}). The absence of a discernible trend in the D5 of CBN indicates that while the Ukiyo-e artists produced less disruptive works in the later period (Figure~\ref{fig:trajectory_cos_D5}
), their efforts were likely directed toward developing the style of each artwork (Figure~\ref{fig:trajectory_cluster_D5}).

\subsection{Quantifying the Influence of Individual Artists in Master-Disciple Relationships}

Our analysis with D5 reveals the trajectory of disruption in ukiyo-e artworks; however, the specific roles of individual artists within these observed trajectories remain unexplored. As discussed in \cite{zeng2023disruptive}, where the roles of scientists in scientific works are examined, each artist in the ukiyo-e tradition serves distinct and significant roles in the creation of art pieces. In ukiyo-e, the master-disciple relationship is crucial, with masters often bestowing their names upon their disciples. This tradition extends beyond names to include the inheritance of styles, techniques, and motifs across generations~\cite{Davis2014LectureNotes}.

The Utagawa school exemplifies this practice, where Utagawa Toyokuni I mentored numerous disciples, including Utagawa Kuniyoshi and Utagawa Hiroshige. These disciples inherited and expanded upon the stylistic elements and thematic choices of their master. For instance, Utagawa Kuniyoshi’s "Takiyasha the Witch and the Skeleton Spectre" showcases his innovation within the established style by integrating dynamic compositions and supernatural themes—an evolution of the dramatic presentation found in Toyokuni’s works. Likewise, in Kuniyoshi’s series depicting warriors and animals, the distinctive influences from his master’s role as an established artist in kabuki portrayals become evident, yet Kuniyoshi’s works adopt a more imaginative and expressive flair.

In contrast, Utagawa Hiroshige’s landscapes, though stylistically connected to the Utagawa school, emphasize serene, atmospheric qualities, demonstrating his individual approach. Through such examples, the significance of the master-disciple relationship in ukiyo-e becomes clear: it is a means by which artistic traditions are both preserved and transformed across generations.

\begin{figure}[!htb]  
  \centering
  \begin{subfigure}{0.495\textwidth}
    \centering
    \includegraphics[width=1.0\linewidth]{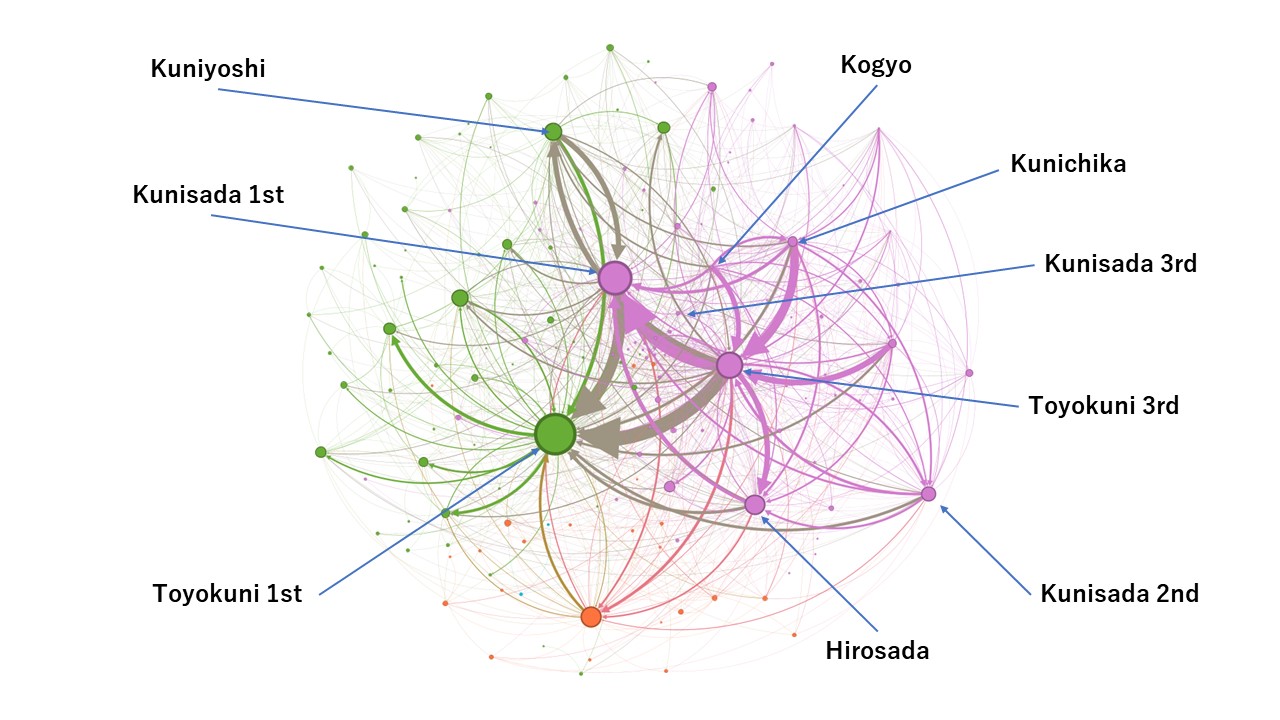}
    \caption{Cosine Similarity Network (CSN)}
    \label{fig:cosine_similarity}
  \end{subfigure}
  \hfill
  \begin{subfigure}{0.495\textwidth}
    \centering
    \includegraphics[width=1.0\linewidth]{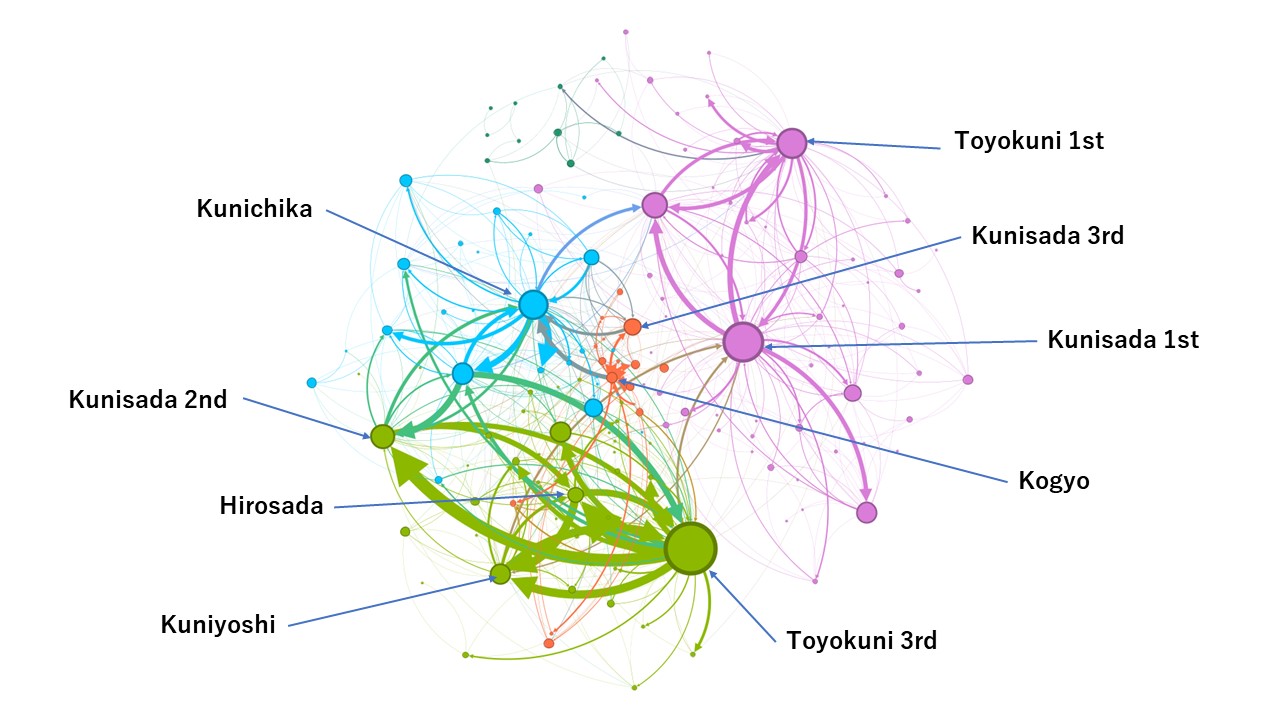}
    \caption{Clustering Based Network (CBN)}
    \label{fig:K-means_clustering}
  \end{subfigure}
\caption{Ukiyo-e artists networks}\label{fig:artist_network}
\begin{minipage}{1\textwidth}
{ The representation of influence among ukiyo-e artists through the aggregation of a network of ukiyo-e works. The size of the arrows indicates the total number of aggregated works and each color represents cluster affiliations based on the modularity-based community detection\cite{blondel2008fast}. Since the CSN network is based on the similarity between works, its aggregation in Figure~\ref{fig:cosine_similarity} reflects the influence of each artist as measured by the similarity of their works over time. On the other hand, the CBN network is constructed based on the style to which the works belong. Therefore, its aggregation in Figure~\ref{fig:K-means_clustering} captures the influence of ukiyo-e artists within the context of their respective style.}
\end{minipage}
\end{figure}

To model the individual roles of artists and their master-disciple relationships, we aggregate the networks of Ukiyo-e artworks discussed in Section~\ref{sec:artwork_disruptiveness} (Method Sec.~\ref{sec:artwork_network}). As we employ the two distinctive network representations CSN and CBN, we generate two corresponding types of individual artist networks where each node represents an artist and each directed edge depicts the influence (Fig.~\ref{fig:artist_network},). The figure reveals the heterogeneous effect of individual artists in the networks, indicating a few groups of artists have significant influences on other artists. We find that the works by those groups of ``influencers'' are referenced by later artists in both types of representation (Fig.\ref{fig:K-means_clustering} and \ref{fig:cosine_similarity}). Figure~\ref{fig:cosine_similarity}, for example, shows that the largest node is Toyokuni 1st, a famous artist known for establishing his own style, with the Toyokuni lineage continuing through a master-disciple relationship up to the seventh generation (who assumed the name in 2001).

To quantitatively assess this observation on the network visualization, we first calculate the betweenness centrality values and presented in Table~\ref{table:cos_sim}. The table shows show that certain artists exhibit higher centrality values, indicating their central roles within the Ukiyo-e art style propagation depicted in CSN. The table also describes that specific artists such as Toyokuni 1st or 3rd appear with the highest betweenness centrality, confirming their significant impact on the evolution of Ukiyo-e art. Since the betweenness centrality measurement of a given artist can significantly be affected by the number of works published by that artist, we calculate disruptiveness scores of individual artists to conduct a fair evaluation of the influence of artists on subsequent generations. Table~\ref{table:cos_artist_D5}  evaluates the disruptiveness of the individual artist calculated on the constructed individual artists networks and shows high values for Toyokuni 1st and First Generation Kunisada 1st. Notably, Generation Toyokuni 3rd, who had the highest score in betweenness centrality, ranks fourth in terms of disruptiveness.

We also calculate the betweenness centrality of CBN, yielding similar results to the CSN (Table~\ref{table:kmean_clus}), where Toyokuni 1st, 3rd, and Kunisada 1st have high scores. However, our analysis with D5, presented in Table~\ref{table:cluster_artist_D5}, produces a different result. First, the artist network of CBN displays a higher average D5 than CSN. We find that Kunisada 1st still scores higher than Kunisada 3rd in their D5, but both attain high scores. This suggests that each Ukiyo-e artist pursued originality in painting style as a means of expressing individuality. The artists covered in this case study are not only prolific but also historically significant, which likely contributed to their establishing a distinctive style, resulting in a generally high and stable level of disruptiveness.

It is notable that some artists, such as Kunisada 1st and Toyokuni 1st, show high D5 scores in both networks. One plausible explanation would be that the visual characteristics and painting style of Ukiyo-e are not entirely separable and independent. Being visually similar and having a similar painting style can occur simultaneously, leaving room for discussion of whether the diffusion of visual images secondarily influenced the painting style or vice versa. Analysis using disruptiveness scores also evaluates the novelty in the painting style.

\section{Discussion and Conclusion}

A comprehensive machine learning-based analysis of Ukiyo-e artworks anchors this study, with a sharp focus on artists' influence and creativity within the Ukiyo-e tradition. Deploying deep convolutional networks for feature extraction and network construction based on these features, the study quantitatively assesses the stylistic evolution and creative impact of individual artists.

Key insights emerged from the analysis regarding the propagation and evolution of Ukiyo-e styles. Betweenness centrality scores highlighted the pivotal roles of certain artists, especially Third Generation Toyokuni, in the stylistic development of Ukiyo-e. Such artists, acting as bridges in the stylistic network, left a significant mark on the evolution of Ukiyo-e art. Variations in the peak periods of betweenness centrality across the CSN and CBN reveal differences in style evolution and its influence on successive works. A notable decrease in the transmission of Ukiyo-e styles around the Meiji Restoration period, marked by a decline in Ukiyo-e production and shifts in betweenness centrality, emphasizes the influence of external cultural and technological factors on the art form.

Employing the D5 disruptiveness score offered a nuanced perspective on Ukiyo-e artists' creativity. Artists like First Generation Toyokuni and Kunisada, with high D5 scores, emerged as pioneers introducing novel and influential styles within the Ukiyo-e tradition. Notably, despite Third Generation Toyokuni's high rank in mediation, his D5 score revealed a propensity for propagating existing styles over creating new ones, distinguishing the diverse roles artists played in the Ukiyo-e tradition—some as innovators, others as carriers of established styles.

This study enhances the understanding of Ukiyo-e through several avenues. We introduce a quantitative method for analyzing artists' stylistic evolution and influence in a field traditionally subjective. Bridging traditional art historical methods with modern computational techniques, it presents a fresh perspective on art history. Furthermore, it underlines machine learning's capacity to reveal art patterns and relationships not easily discerned through conventional analysis.

Nevertheless, the study acknowledges its limitations, such as potentially overlooking the contributions of less prolific artists due to focusing on those represented by over 100 images in the dataset. Moreover, while powerful, the computational approach might not fully grasp the depth and nuance of artistic creativity and influence, shaped by a myriad of cultural, historical, and personal factors.

In essence, the result underscores the potential of machine learning in art historical research, opening new avenues for understanding the evolution and creativity of Ukiyo-e art and beyond. It sets the stage for further research across various art forms, applying similar computational techniques to explore artistic influence and evolution.

\appendix
\section*{Appendix} 
\renewcommand{\thetable}{\Alph{section}.\arabic{table}} 

\begin{table}[htbp]
\centering 
\caption{Betweenness Centrality in Cosine Similarity Network (CSN) of Ukiyo-e Artists}
\label{table:cos_sim}
\begin{tabular}{clll}
\textbf{ID} & \textbf{Ukiyo-e Artist (Active Period)} & \textbf{Betweenness Centrality} &  \textbf{Generation (if applicable)}\\
\hline\hline
1 & Kunisada (1808--1843) & 2878.40 & 1st \\
2 &  Kunisada (1851--1870) & 1404.12 & 2nd\\
3 &  Kunisada (1872--1903) & 384.38 & 3rd\\
4 &  Toyokuni (1793--1827) & 3768.65 & 1st\\
5 &  Toyokuni (1844--1864) & 5647.75 & 3rd\\
6 & Kuniyoshi (1825--1855) & 1302.76 & \\
7 & Kogyo (1897--1924) & 28.79 & \\
8 & Kunichika (1862--1899) & 1440.83 & 	\\
9 & Hirosada (1847--1855) & 874.84 & 	
\end{tabular}
\end{table}

\begin{table}[htbp]
\centering 
\caption{Betweenness Centrality in K-means Clustering Network (CBN) of Ukiyo-e Artists}
\label{table:kmean_clus}
\begin{tabular}{clll}
\textbf{ID} & \textbf{Ukiyo-e Artist (Active Period in Year)} & \textbf{Betweenness Centrality} & \textbf{Generation (if applicable)}\\

\hline\hline
1 & Kunisada (1808--1843) & 4236.89 & 1st\\
2 & Kunisada (1851--1870) & 616.60 & 2nd\\
3 & Kunisada (1872--1903) & 445.65 & 3rd\\
4 & Toyokuni (1793--1827) & 3708.73 & 1st\\
5 & Toyokuni (1844--1864) & 5613.47 & 3rd\\
6 & Kuniyoshi (1825--1855) & 1131.62\\
7 & Kogyo (1897--1924) & 837.61\\
8 & Kunichika (1862--1899) & 3785.30\\
9 & Hirosada (1847--1855) & 398.02
\end{tabular}
\end{table}

\begin{table}[htbp]
\centering 
\caption{D5 in cos similarity networks (CSN) of ukiyo-e artists}
\label{table:cos_artist_D5}
\begin{tabular}{clll}
\textbf{ID} & \textbf{Ukiyo-e Artist (Active Period in Year)} & \textbf{D5} & \textbf{Generation (if applicable)}\\

\hline\hline
1 & Kunisada (1808--1843) & 0.71 & 1st\\
2 & Kunisada (1851--1870) & 0.56 & 2nd\\
3 & Kunisada (1872--1903) & 0.36 & 3rd\\
4 & Toyokuni (1793--1827) & 0.72 & 1st\\
5 & Toyokuni (1844--1864) & 0.63 & 3rd\\
6 & Kuniyoshi (1825--1855) & 0.65\\
7 & Kogyo (1897--1924) & 0.30\\
8 & Kunichika (1862--1899) & 0.46\\
9 & Hirosada (1847--1855) & 0.58
\end{tabular}
\end{table}

\begin{table}[htbp]
\centering 
\caption{D5 in clustering networks (CBN) of ukiyo-e artists}
\label{table:cluster_artist_D5}
\begin{tabular}{clll}
\textbf{ID} & \textbf{Ukiyo-e Artist (Active Period in Year)} & \textbf{D5} & \textbf{Generation (if applicable)}\\

\hline\hline

1 & Kunisada (1808--1843) & 0.92 & 1st\\
2 & Kunisada (1851--1870) & 0.90 & 2nd\\
3 & Kunisada (1872--1903) & 0.82 & 3rd\\
4 & Toyokuni (1793--1827) & 0.94 & 1st\\
5 & Toyokuni (1844--1864) & 0.91 & 3rd\\
6 & Kuniyoshi (1825--1855) & 0.92& \\
7 & Kogyo (1897--1924) & 0.90& \\
8 & Kunichika (1862--1899) & 0.92& \\
9 & Hirosada (1847--1855) & 0.88& 

\end{tabular}
\end{table}

\end{document}